%% file: acl_latex.tex
\pdfoutput=1
\documentclass[11pt]{article}

\usepackage[preprint]{acl}
\usepackage{fontawesome}
\usepackage{times}
\usepackage{latexsym}
\usepackage{balance}

\usepackage[T1]{fontenc}

\usepackage[utf8]{inputenc}

\usepackage{microtype}

\usepackage{inconsolata}

\usepackage{graphicx}
\usepackage{pgfplots}
\pgfplotsset{compat=1.18}
\usetikzlibrary{arrows.meta,positioning,fit,backgrounds,calc}
\usepackage{xcolor}
\usepackage{enumitem}
\usepackage{booktabs}
\usepackage{multirow}

\usepackage{siunitx}
\title{EviSearch: Trustworthy Extraction and Synthesis of Clinical Trial Evidence with Agents that Improve with Use}

\author{
\textsuperscript{1}Naman Ahuja  \quad
\textsuperscript{1}Abhijit Chakraborty \quad 
\textbf{\textsuperscript{2}Muhammad Ali Khan} \quad \\
\textbf{\textsuperscript{2}Kaneez Zahra Rubab Khakwani} \quad 
\textbf{\textsuperscript{2}Mohamad Bassam Sonbol} 
\textbf{\textsuperscript{2}Irbaz Bin Riaz} \quad
\textbf{\textsuperscript{1}Vivek Gupta} \\
Arizona State University\textsuperscript{1} \quad
Mayo Clinic\textsuperscript{2} \\
\faGithub \ \href{https://github.com/CoRAL-ASU/EviSearch}{Code} \quad 
\faVideoCamera \ \href{https://evisearch.fly.dev/}{Demo} \quad
\faPlay \ \href{https://drive.google.com/file/d/1JoYi5WpjAJUtDApLi0EC4lc2z3vPvu2Z/view?usp=sharing}{Video} \\
\texttt{riaz.irbaz@mayo.edu, vgupt140@asu.edu}
}

\input{sections/numbers}

\begin{document}
\maketitle

\input{sections/abstract}

\begin{figure*}[t!]
\centering
\resizebox{\textwidth}{!}{\input{figures/architecture}}
\caption{Architecture of EviSearch. Three agents with complementary access to the publication extract every column;
the Reconciliation Agent extracts independently before adjudicating the three candidates. A value is admitted only
after the attribution verifier has read it on its cited page, and cells on which the extraction agents disagree are
routed to review, whose corrections refine the schema and the curation knowledge base. All model inference within
the dotted boundary runs on open-weight models locally.}
\label{fig:system}
\end{figure*}
\input{sections/intro}

\input{sections/related}
\input{sections/system}
\input{sections/evaluation}

\input{sections/demo}
\input{sections/conclusion}
\input{sections/limitations}

\section*{Acknowledgements}
This research was supported by the Mayo Clinic
and Arizona State University Alliance for Health
Care Collaborative Research Seed Grant Program (Award ID: AWD00041508; Sponsor Award
ID: ARI-358187) for the project: ‘Artificial
intelligence-assisted digital, living, interactive clinical practice guidelines for cancer providers and
patients.’

\input{sections/ethics}

\balance
\bibliography{custom}

\appendix
\input{sections/appendix}

\end{document}

%% file: sections/numbers.tex
\newcommand{\nColumns}{133}
\newcommand{\batchSize}{15}

\newcommand{\accBaselineRunOne}{88.93}
\newcommand{\accBaselineRunTwo}{89.10}
\newcommand{\accBaseline}{89.01}
\newcommand{\accAgentA}{89.71}
\newcommand{\accAgentB}{92.27}
\newcommand{\accOracleAB}{93.21}
\newcommand{\accSystem}{91.70}
\newcommand{\accSystemRunOne}{91.88}
\newcommand{\accSystemRunTwo}{91.52}
\newcommand{\accAgentARunOne}{89.42}   \newcommand{\accAgentARunTwo}{90.00}
\newcommand{\accAgentBRunOne}{92.16}   \newcommand{\accAgentBRunTwo}{92.37}
\newcommand{\accOracleRunOne}{93.20}   \newcommand{\accOracleRunTwo}{93.23}
\newcommand{\gapToAgentB}{0.57}
\newcommand{\gainOverBaseline}{2.69}

\newcommand{\citedNow}{100.0}        
\newcommand{\citedBaseline}{0.0} \newcommand{\citedAgentA}{99.7} \newcommand{\citedAgentB}{100.0}
\newcommand{\citedOracle}{99.9}
\newcommand{\verifiedNone}{0.0}      
\newcommand{\corroborated}{98.9}     
\newcommand{\highlighted}{99.1}      
\newcommand{\valueBoxed}{73}         
\newcommand{\valuePartBoxed}{9}      
\newcommand{\gateDropCorrectShare}{53}   
\newcommand{\agreeDropped}{2.1}   \newcommand{\provDropped}{1.1}
\newcommand{\agreeDropCorrect}{53} \newcommand{\provDropCorrect}{50}   
\newcommand{\agreeFixed}{0.26}   \newcommand{\provFixed}{0.38}
\newcommand{\agreeBroke}{0.34}   \newcommand{\provBroke}{0.34}
\newcommand{\refAuditShare}{0.8}   

\newcommand{\queueFrac}{15.6}        
\newcommand{\queueAcc}{95.22}        
\newcommand{\queueAccRunOne}{95.09} \newcommand{\queueAccRunTwo}{95.34}
\newcommand{\queueRandomSys}{93.00}  
\newcommand{\queueRandomB}{93.47}    
\newcommand{\queueGainOverRandomB}{1.75}
\newcommand{\queueRecall}{40}        
\newcommand{\queuePrecision}{27}     
\newcommand{\tierTwoFrac}{63.4}
\newcommand{\tierTwoAcc}{98.04}

\newcommand{\draftA}{88.35}   \newcommand{\draftB}{90.86}    
\newcommand{\revA}{89.31}     \newcommand{\revB}{91.80}      
\newcommand{\revfourA}{89.44} \newcommand{\revfourB}{91.23}  
\newcommand{\kbA}{89.71}      \newcommand{\kbB}{92.27}       
\newcommand{\draftSys}{89.91} \newcommand{\revSys}{90.49} \newcommand{\revfourSys}{91.04}
\newcommand{\kbSys}{91.70} \newcommand{\kbGainSys}{0.66} \newcommand{\totalGainSys}{1.79}
\newcommand{\kbGainA}{0.27}   \newcommand{\kbGainB}{1.04}    
\newcommand{\totalGainA}{1.36} \newcommand{\totalGainB}{1.41} 

\newcommand{\modelName}{Qwen3.6-27B}
\newcommand{\embedName}{Qwen3-Embedding-8B}
\newcommand{\minutesPerPaper}{5}

%% file: sections/abstract.tex
\begin{abstract}
Structured extraction of evidence from clinical trial publications underpins systematic reviews and clinical
guidelines, yet large language models are adopted for it only hesitantly: their outputs are difficult to verify,
their use commonly requires transmitting documents to proprietary services, and they do not improve from the
corrections their users make. We present \textbf{EviSearch}, a multi-agent system that addresses these three
obstacles. Three tool-augmented agents with complementary access to a publication extract every column of an evidence
table, and a value is admitted only after an attribution verifier has read it on its cited page, so that every value
carries a page-level attribution. Disagreement between independent agents directs human review to the cells most
likely to be wrong, and reviewer feedback refines the schema definitions and a curation knowledge base without
updating model parameters. The agentic system runs entirely offline on open-weight models. On a clinician-annotated
benchmark of randomized-trial publications, EviSearch attributes \citedNow\% of its values, reaches \accSystem\% accuracy
autonomously, and reaches \queueAcc\% after review of \queueFrac\% of cells, exceeding random review of the strongest
single agent at equal effort by \queueGainOverRandomB{} points.
\end{abstract}

%% file: figures/architecture.tex
\begin{tikzpicture}[
    font=\footnotesize,
    box/.style      ={draw, rounded corners=2pt, align=center, inner sep=3.5pt, minimum height=8mm},
    srcbox/.style   ={box, fill=black!5},
    agentbox/.style ={box, fill=blue!7,    minimum width=33mm},
    judgebox/.style ={box, fill=blue!13,   minimum width=33mm},
    verbox/.style   ={box, fill=orange!13, minimum width=31mm},
    gatebox/.style  ={box, fill=orange!26, minimum width=22mm},
    outbox/.style   ={box, fill=green!10,  minimum width=30mm},
    arr/.style      ={-{Stealth[length=1.7mm]}, semithick},
    fb/.style       ={-{Stealth[length=1.7mm]}, semithick, densely dashed},
    lbl/.style      ={font=\scriptsize, inner sep=1.2pt},
  ]

  \node[srcbox] (pdf) at (0,0.9) {Trial publication\\[-1pt](PDF)};
  \node[srcbox] (parse) at (0,-0.9) {Layout parsing\\[-1pt]\scriptsize page text $\cdot$ page images\\[-2pt]\scriptsize $\cdot$ retrieval index};

  \node[agentbox] (a) at (4.3,1.45)
       {\textbf{PDF Query Agent (A)}\\[-1pt]\scriptsize full document: text + page images\\[-2pt]\scriptsize single structured call};
  \node[agentbox] (b) at (4.3,0)
       {\textbf{Search Agent (B)}\\[-1pt]\scriptsize dense retrieval over page text\\[-2pt]\scriptsize tool-augmented loop};
  \node[agentbox] (c) at (4.3,-1.45)
       {\textbf{Reconciliation Agent}\\[-1pt]\scriptsize independent extraction phase\\[-2pt]\scriptsize retrieval + page images};

  \node[judgebox] (adj) at (8.6,0)
       {\textbf{Reconciliation Agent}\\[-1pt]\scriptsize adjudication phase\\[-2pt]\scriptsize three candidates $\rightarrow$ one value};
  \node[verbox] (ver) at (8.6,-1.95)
       {\textbf{Attribution verifier}\\[-1pt]\scriptsize reads the cited pages and\\[-2pt]\scriptsize answers independently};

  \node[gatebox] (gate) at (12.1,0) {\textbf{Provenance-}\\[-1pt]\textbf{gated}\\[-1pt]\textbf{admission}};

  \node[outbox] (table) at (15.6,1.35) {\textbf{Evidence table}\\[-1pt]\scriptsize value, page, quotation};
  \node[outbox] (queue) at (15.6,0)    {\textbf{Review triage}\\[-1pt]\scriptsize cells where A and B disagree};
  \node[outbox] (kb)    at (15.6,-1.45) {\textbf{Curation knowledge}\\[-1pt]\scriptsize schema definitions + notes};

  \begin{scope}[on background layer]
    \node[draw=black!55, densely dotted, rounded corners=4pt, fill=black!2, inner sep=7pt,
          fit=(a) (b) (c) (adj) (ver) (gate), label={[lbl, anchor=south west, xshift=2pt]north west:%
          \textit{Agentic system: open-weight models, local inference, fully offline}}] (offline) {};
  \end{scope}

  \draw[arr] (pdf) -- (parse);
  \draw[arr] (parse.east) -- ++(0.45,0) |- (a.west);
  \draw[arr] (parse.east) -- ++(0.45,0) |- (b.west);
  \draw[arr] (parse.east) -- ++(0.45,0) |- (c.west);

  \draw[arr] (a.east) -| ($(adj.west)+(-0.4,0)$) -- (adj.west);
  \draw[arr] (b.east) -- (adj.west);
  \draw[arr] (c.east) -| ($(adj.west)+(-0.4,0)$) -- (adj.west);

  \draw[arr] ([xshift=6mm]adj.south) -- node[lbl, right] {value} ([xshift=6mm]ver.north);
  \draw[arr] ([xshift=-6mm]ver.north) -- node[lbl, left, align=right] {attribution} ([xshift=-6mm]adj.south);

  \draw[arr] (adj) -- (gate);
  \draw[arr] (gate.east) -- ++(0.35,0) |- (table.west);
  \draw[arr] (gate.east) -- ++(0.35,0) |- node[lbl, above, pos=0.8] {A $\neq$ B} (queue.west);
  \draw[arr] (queue) -- node[lbl, right] {corrections} (kb);

  \draw[fb] (kb.south) -- ++(0,-1.05) -| node[lbl, below, pos=0.27] {schema alignment: definitions and notes delivered to every agent} (c.south);

\end{tikzpicture}

%% file: sections/intro.tex
\section{Introduction}
\label{sec:intro}

Systematic reviews and clinical practice guidelines rest on structured data extracted from the publications of
clinical trials. The extraction is laborious, contributing to the more than a year that a systematic review typically
requires~\cite{borah2017analysis}, and it is error-prone: data extraction could not be reproduced for 17\% of trials in
one large audit of meta-analyses~\cite{xu2022validity}, and extraction error rates of up to 50\% have been
reported~\cite{mathes2017frequency}. Living systematic reviews, which are updated as new trials appear, turn this cost
into a recurring one~\cite{elliott2014living}.

Large language models (LLMs) can assist with data extraction~\cite{khraisha2024can,konet2024performance}, and
systems built on them have accelerated evidence synthesis under expert supervision~\cite{wang2025accelerating}. Three
obstacles nonetheless limit their adoption in curation workflows. The first is \emph{verifiability}. LLM extractions
contain errors and hallucinations~\cite{huang2023survey,gartlehner2025from}, current tools are regarded as aids to
rather than replacements for human reviewers~\cite{yisha2026assessing}, and clinicians require that model outputs be
checkable before they are relied upon~\cite{tonekaboni2019what,hager2024evaluation}. An extracted value without an
attribution to its source can be verified only by re-reading the publication. The second is \emph{data governance}.
Commercial LLM services require documents to be transmitted to remote servers, which raises legal and ethical
concerns for clinical data~\cite{wiest2024privacypreserving,marks2023chatbots}, whereas open-weight models deployed
locally are increasingly competitive on clinical information extraction~\cite{nowak2025privacyensuring,dorfner2024comparing}
and are reproducible by construction~\cite{riedemann2024path}. The third is that extraction systems are
\emph{static}: the corrections that curators make to one table do not inform the next.

We present \textbf{EviSearch}, a multi-agent system for clinical evidence extraction designed around these
obstacles. Three tool-augmented agents with complementary access to a publication extract every column of an evidence
table. A value is admitted only after an attribution verifier has read it on its cited page, so that every value in
the table carries a page, a supporting quotation, and a modality. Building on the observation that disagreement
between independent LLM extractions identifies inaccurate responses~\cite{khan2025collaborative}, EviSearch routes the
cells on which its agents disagree to human review. Reviewer feedback refines the schema definitions and a curation
knowledge base that every agent reads, so the system improves with use without updating model parameters. The agentic
system runs entirely offline on open-weight models. Our contributions are as follows:
\begin{itemize}[leftmargin=*,itemsep=1pt,topsep=2pt]
  \item \emph{Provenance-gated admission}, under which every admitted value carries an attribution read by an
    independent verifier, and \corroborated\% of values are independently reproduced.
  \item \emph{Disagreement-based review triage}, which raises accuracy from \accSystem\% to \queueAcc\% by directing
    review to \queueFrac\% of cells, \queueGainOverRandomB{} points above random review of the strongest single agent
    at equal effort.
  \item \emph{Improvement with use} through schema alignment and a traceable curation knowledge base.
  \item An open-source, fully offline implementation with a web interface for defining tables, inspecting
    attributions, and reviewing extractions.
\end{itemize}

%% file: sections/related.tex
\section{Related Work}
\label{sec:related}

\paragraph{LLMs for evidence synthesis.} LLMs have been evaluated for data extraction in systematic
reviews~\cite{khraisha2024can,konet2024performance,gartlehner2025from}, and TrialMind integrates them into a
supervised synthesis workflow~\cite{wang2025accelerating}. \citet{khan2025collaborative} showed that concordance
between two LLMs separates accurate from inaccurate extractions in living systematic reviews. These studies
primarily report extraction accuracy; EviSearch additionally attributes and verifies every value and turns
inter-agent disagreement into an operational review queue.

\paragraph{Attribution and verification.} Attributing generated text to its sources is the basis of verifiable
generation~\cite{rashkin2023measuring,gao2023enabling}, yet citations produced by generative systems are frequently
incomplete or unsupported~\cite{liu2023evaluating}. Verification by a separate step or model improves factual
reliability~\cite{dhuliawala2024chain,tang2024minicheck}, and retrieval grounds model outputs in source
documents~\cite{lewis2020rag,xiong2024benchmarking}. Tool-augmented agents interleave retrieval with
reasoning~\cite{yao2023react,trivedi2023interleaving}, including over scientific literature~\cite{lala2023paperqa}.
EviSearch conditions admission to its output on an independent reading of the cited page.

\paragraph{Uncertainty and human--AI triage.} Inconsistency among sampled or independent outputs signals
error~\cite{wang2023selfconsistency,manakul2023selfcheckgpt,farquhar2024detecting}, and uncertainty can allocate items
between models and human annotators~\cite{li2023coannotating,wang2024humanllm} or defer clinical cases to
experts~\cite{dvijotham2023enhancing}. EviSearch derives its triage signal from agents with complementary access to
the document rather than from repeated samples of one model.

\paragraph{Learning without parameter updates.} LLM systems can improve from feedback through external memory
and accumulated experience~\cite{madaan2022memory,dalvimishra2022towards,zhao2024expel}, and annotation guidelines
improve information extraction~\cite{sainz2024gollie,pang2023guideline}. EviSearch maintains such knowledge as
human-curated documents whose every revision is traceable to the extraction runs that used it.

%% file: sections/system.tex
\section{EviSearch}
\label{sec:system}

EviSearch converts a clinical trial publication into a row of a structured evidence table in which every value is
attributed to the passage from which it was extracted (Figure~\ref{fig:system}). The system comprises three
tool-augmented agents with complementary access to the document, an attribution verifier that governs admission to
the table, and a curation layer through which reviewer feedback refines the behaviour of every agent. All model
inference is performed by open-weight models on local hardware.

\subsection{Task Formulation}
\label{sec:task}

Given a publication $D$ and a schema $\mathcal{S} = \{(c_i, d_i)\}_{i=1}^{n}$ of $n$ columns with natural-language
definitions $d_i$, the system produces for each column either a value $v_i$ with an attribution
$a_i = (p_i, q_i, m_i)$, comprising the source page, a supporting quotation, and the evidence modality (text, table, or
figure), or the abstention label \textsc{NR}. The schema used here contains \nColumns{} columns drawn from a living
evidence platform for metastatic castration-sensitive prostate cancer, covering trial design, treatment arms,
baseline characteristics, efficacy outcomes, subgroup results, and adverse events. Each definition specifies the
statistic, population, arm, timepoint, and unit the column requires, and the conventions that apply when a
publication reports the quantity under another name or not at all. Columns are extracted in batches of at most
\batchSize{}, grouped by clinical section; every agent works over the same batches, making their outputs comparable
cell by cell.

At ingestion, each publication is parsed once into page-level text with modality labels, its pages are rendered to
images locally, and the page text is embedded with \embedName{} into a per-document dense index. Layout parsing is
the only operation that calls an external service, once per document.

\subsection{Extraction Agents}
\label{sec:agents}

\paragraph{PDF Query Agent (Agent A).} The PDF Query Agent receives the complete document, namely the parsed text
followed by every rendered page image, and answers a batch in a single structured-output call without tools. With the
entire publication in context, it is not limited by retrieval recall and suits columns that describe the document as
a whole, such as trial design or whether an outcome is reported.

\paragraph{Search Agent (Agent B).} The Search Agent operates without an initial document context and acquires
evidence exclusively through retrieval. It has two tools: \texttt{search\_chunks}, dense retrieval over the
page-level index returning the full text of the highest-ranked pages, and \texttt{get\_chunks\_by\_page}, which
retrieves pages by number. It follows an extract-first policy, answering the columns the opening pages support before
retrieving for the remainder, and deduplicates pages already in context, bounding context growth over long batches.
Because it consumes parsed text only, values rendered solely within figures lie outside its observable evidence.

\subsection{Reconciliation and Provenance-Gated Admission}
\label{sec:reconcile}

The Reconciliation Agent adjudicates the outputs of the two extraction agents through a two-phase protocol. In the
\emph{independent extraction phase}, the candidate outputs are withheld and the agent extracts the batch itself, using
the Search Agent's retrieval tool together with \texttt{get\_pages}, which returns both the parsed text and the
rendered image of a page. This ordering is enforced by the available tool set rather than by instruction. In the
\emph{adjudication phase}, the candidate values of Agents A and B are revealed together with their reasoning and cited
evidence. The agent determines a final value for each column and records a verdict (correct, incomplete, incorrect,
or no answer) for each of the three candidates, its own included.

\paragraph{Attribution verifier.} Given a candidate value and the pages it cites, the verifier reads the page text
and rendered image and must formulate its own answer for the column before comparing it with the candidate; it
returns the page, a supporting quotation, the modality, and whether its reading reproduces the candidate. Answering
first guards against confirming a value merely because it appears on the page: a number printed there may answer a
different question than the column asks.

\paragraph{Provenance-gated admission.} A value is admitted to the table only after it has been read on its cited
page by the attribution verifier, which supplies the attribution that the interface presents. Admission is
conditioned on provenance rather than on agreement: when the verifier's independent reading does not reproduce a
value, the value is admitted with the quotation cited by its extracting agent, marked as uncorroborated, and routed to
review. An abstention is admitted only with an explicit basis, stating the pages read and what they report for the
column. Every admitted value therefore carries an attribution, and the corroboration mark distinguishes the values an
independent reading reproduced from the rest.

\paragraph{Evidence location.} An attribution names a page; the reviewer needs the region of it that holds the
value. EviSearch locates regions on the cited pages in order of specificity: the value as printed, taking among
repeated occurrences the one nearest the supporting quotation and the column's terms; for a derived value, the
numbers it was computed from, read from the arithmetic stated in its reasoning; and otherwise the supporting passage.
Each region is stored with its page coordinates, so the interface highlights the value itself rather than the
paragraph around it.

\subsection{Disagreement-Based Review Triage}
\label{sec:triage}

Agents A and B differ in how they access a document: one observes the entire publication including its figures,
the other only the pages its retrieval selects, as text. Their errors are consequently only partially correlated, and
the cells on which they disagree concentrate the errors of the table. EviSearch routes these cells, together with
values the attribution verifier did not corroborate, to a review queue, which shows each cell's three candidates, the
verdicts recorded by the Reconciliation Agent, the verifier's reading, and the cited region highlighted in the
source.

\subsection{Improvement with Use}
\label{sec:learning}

EviSearch is aligned to its user through two channels, neither of which updates model parameters.

\paragraph{Schema alignment.} The schema is drafted automatically from the column headers of the target table and a
single example row. The drafting agent attaches clarifying questions to definitions whose intent is ambiguous; the
table owner accepts, edits, or answers them; and the agent revises the affected definitions before the schema is
locked as a new version.

\paragraph{Knowledge Base Curation.} Conventions that extend beyond individual definitions are maintained as
knowledge notes: Markdown documents, each covering one topic with its scope (global, a family of columns, or
specific columns) and its exceptions. \emph{Definition} notes, which state what a column means, are delivered to
every agent; \emph{extraction} notes, which state how values are located and assembled, are withheld from the
Reconciliation Agent's independent extraction, so that it does not inherit the procedure of the agents it
adjudicates. A reviewer correction that reflects an interpretation of the schema, rather than a misreading of a
single document, becomes an edit of the note governing the column, logged with the reviewer, its rationale, and the
resulting content fingerprint of the knowledge base. Every extraction run records that fingerprint, so any value can
be traced to the knowledge in force when it was produced.

\subsection{Deployment}
\label{sec:deploy}

All three agents and the attribution verifier run on \modelName{}~\cite{yang2025qwen3} and retrieval uses
\embedName{}~\cite{zhang2025qwen3embedding}, both open-weight. Served with vLLM~\cite{kwon2023vllm} on a single
NVIDIA H200 GPU, the agentic system operates entirely offline: no document content is transmitted to an external model
provider during extraction, verification, or adjudication. The same models are available from hosted providers, which
the public demonstration uses. Execution follows the data dependencies alone: the extraction agents never read each
other's output and every column batch is an independent session, so both agents and all of their batches run
concurrently, after which the Reconciliation Agent processes its batches concurrently. Served by a hosted provider, a
publication is processed end to end in about \minutesPerPaper{} minutes.

%% file: sections/evaluation.tex
\section{Evaluation}
\label{sec:eval}

We evaluate the four properties EviSearch is designed to provide: autonomous accuracy, traceability, the reviewer
effort required to reach a given accuracy, and the effect of alignment through use.

\subsection{Setup}
\label{sec:setup}

\paragraph{Benchmark.} The benchmark comprises publications of randomized controlled trials in metastatic
castration-sensitive prostate cancer, each annotated by clinicians with reference values for all \nColumns{} columns,
as released (Appendix~\ref{app:eval}).

\paragraph{Scoring.} Each value is scored against its reference by Claude Opus~5 as an LLM
judge~\cite{zheng2023judging,croxford2025evaluating}, blind to the producing system, under a written rubric that
distinguishes exact-match, numeric-tolerance, and structured-text columns and fixes when two values are equivalent
(Appendix~\ref{app:eval}). A cell is rated for correctness and completeness on $\{0, 0.5, 1\}$; its score is the mean
of the two, and accuracy is the mean over all cells. Every configuration was run twice.

\paragraph{Comparisons.} The \emph{single-pass} baseline extracts each batch in one call from the parsed text with
the same model, schema, and knowledge as EviSearch. We also report each extraction agent in isolation and
\emph{oracle selection}, which takes the better of the two agents' values per cell and so bounds any method that
selects between them.

\begin{table}[t]
\centering
\small
\setlength{\tabcolsep}{2.3pt}
\begin{tabular}{@{}lccccc@{}}
\toprule
 & \multicolumn{3}{c}{\textbf{Accuracy (\%)}} & \multicolumn{2}{c}{\textbf{Attrib.\ (\%)}} \\
\cmidrule(lr){2-4}\cmidrule(l){5-6}
\textbf{System} & Run 1 & Run 2 & Mean & Cited & Verified \\
\midrule
Single pass & \accBaselineRunOne & \accBaselineRunTwo & \accBaseline & \citedBaseline & \verifiedNone \\
PDF Query Agent (A) & \accAgentARunOne & \accAgentARunTwo & \accAgentA & \citedAgentA & \verifiedNone \\
Search Agent (B) & \accAgentBRunOne & \accAgentBRunTwo & \accAgentB & \citedAgentB & \verifiedNone \\
Oracle selection & \accOracleRunOne & \accOracleRunTwo & \accOracleAB & \citedOracle & \verifiedNone \\
\midrule
EviSearch & \accSystemRunOne & \accSystemRunTwo & \accSystem & \citedNow & \corroborated \\
\quad + review of \queueFrac\% & \queueAccRunOne & \queueAccRunTwo & \textbf{\queueAcc} & \citedNow & \corroborated \\
\bottomrule
\end{tabular}
\caption{Extraction accuracy; the share of stated values citing a page and a quotation (Cited), and the share a
second reading of that page reproduces (Verified). Only EviSearch reads each value a second time. Oracle selection
takes the better of Agents A and B per cell; the last row assumes the reviewer corrects every erroneous queued cell.}
\label{tab:accuracy}
\end{table}

\subsection{Extraction Accuracy}
\label{sec:accuracy}

Table~\ref{tab:accuracy} reports autonomous accuracy. EviSearch reaches \accSystem\%, an improvement of
\gainOverBaseline{} points over single-pass extraction with the same model and knowledge. The Search Agent alone is
the most accurate individual configuration, and EviSearch remains within \gapToAgentB{} points of it; neither
approaches oracle selection at \accOracleAB\%. Autonomous accuracy is therefore not the principal contribution: what
distinguishes EviSearch is that every value is attributed and verified, and that the disagreement between its agents
identifies where the remaining errors lie.

\subsection{Traceability}
\label{sec:traceability}

Every value that EviSearch admits carries a page, a supporting quotation, and a modality (\citedNow\%), and for
\corroborated\% of admitted values the attribution verifier's independent reading reproduces the value. The remaining
values are admitted with the quotation cited by their extracting agent and marked for review, so the corroboration
mark conveys information to the reviewer rather than holding by construction. Evidence location resolves
\highlighted\% of admitted values to a highlighted region; for \valueBoxed\% of numeric values every number of the
value lies inside the highlight and for a further \valuePartBoxed\% part of it does, the remainder being values not
printed on their cited page. Conditioning admission on provenance rather than on agreement is deliberate: in an
agreement-gated variant, which admits a value only when the verifier reproduces it, \gateDropCorrectShare\% of the
values removed by its rejections were correct (Appendix~\ref{app:ablation}), because a verifier that fails to locate a
value cannot distinguish its own omission from the absence of the value.

\begin{figure}[t]
\centering
\input{figures/effort}
\caption{Accuracy after review against the fraction of cells reviewed. Triage takes the disagreed or uncorroborated
cells first, then those both extraction agents left empty. Random review of a fraction $f$ recovers a fraction $f$ of
the errors in expectation.}
\label{fig:effort}
\end{figure}

\subsection{Reviewer Effort}
\label{sec:effort}

The triage queue contains \queueFrac\% of all cells, holding \queueRecall\% of the table's errors at a precision
of \queuePrecision\%. Reviewing them raises accuracy from \accSystem\% to \queueAcc\% (Figure~\ref{fig:effort}).
Reviewing the same fraction at random raises EviSearch to \queueRandomSys\% and the most accurate single agent to
\queueRandomB\%: at equal effort, disagreement-based triage improves on the strongest single-agent configuration by
\queueGainOverRandomB{} points and exceeds oracle selection, a signal no single agent provides. Extending review to
the cells both extraction agents left empty reaches \tierTwoAcc\% at \tierTwoFrac\% of cells, beyond which the
marginal return of review approaches that of reading the table exhaustively.

\begin{table}[t]
\centering
\small
\setlength{\tabcolsep}{2.5pt}
\begin{tabular}{@{}llccc@{}}
\toprule
\textbf{Schema} & \textbf{Knowledge} & \textbf{Agent A} & \textbf{Agent B} & \textbf{EviSearch} \\
\midrule
Auto-drafted & guidelines & \draftA & \draftB & \draftSys \\
Reviewed $\times$2 & guidelines & \revA & \revB & \revSys \\
Reviewed $\times$3 & guidelines & \revfourA & \revfourB & \revfourSys \\
Reviewed $\times$3 & + knowledge & \kbA & \kbB & \kbSys \\
\bottomrule
\end{tabular}
\caption{Accuracy (\%) as the schema and the curation knowledge base are aligned through review ($\times$2,
$\times$3: rounds of schema review). \emph{Guidelines} denotes the fixed extraction guidelines without curated
knowledge. Means of two runs.}
\label{tab:alignment}
\end{table}

\subsection{Improvement with Use}
\label{sec:alignment}

Table~\ref{tab:alignment} traces the extraction agents as the schema and the knowledge base are aligned. Review
rounds were conducted by one of the authors: A reviewer with access to the publications, the draft
definitions, and one example row, but never to the reference annotations or to any score. With the extraction
guidelines held fixed, two rounds of schema review improve the PDF Query Agent from \draftA\% to \revA\%, the Search
Agent from \draftB\% to \revB\%, and the reconciled output from \draftSys\% to \revSys\%. The definitions are thus the
primary channel of alignment: they determine what each agent treats as the correct answer. The last two rows hold the
schema fixed at its final version and isolate the curation knowledge base, which improves the Search Agent by
\kbGainB{} points, the PDF Query Agent by \kbGainA{}, and EviSearch by \kbGainSys{}. The larger gain for the Search
Agent is consistent with the role of the extraction notes, which state where values are located and how they are
assembled, and so bear most directly on an agent that must retrieve its evidence. Over the full course of alignment,
the two agents improve by \totalGainA{} and \totalGainB{} points and EviSearch by \totalGainSys{}.

%% file: figures/effort.tex
\begin{tikzpicture}
\begin{axis}[
    width=\linewidth, height=5.0cm,
    xmin=0, xmax=100, ymin=91, ymax=100.4,
    xlabel={Cells reviewed (\%)}, ylabel={Accuracy (\%)},
    xlabel style={font=\footnotesize}, ylabel style={font=\footnotesize},
    tick label style={font=\scriptsize},
    xtick={0,15.6,40,63.4,100}, xticklabels={0,15.6,40,63.4,100},
    ytick={92,94,96,98,100},
    grid=major, grid style={gray!18},
    legend style={font=\scriptsize, at={(0.5,-0.36)}, anchor=north, legend columns=2, draw=none,
                  column sep=6pt, row sep=-1pt, inner sep=1pt},
    legend cell align=left,
  ]
  \addplot[very thick, blue!70!black, mark=*, mark size=1.6pt]
    coordinates {(0,91.70) (15.6,95.22) (63.4,98.04) (100,100)};
  \addlegendentry{EviSearch, triage}
  \addplot[thick, blue!45, dashed]
    coordinates {(0,91.70) (100,100)};
  \addlegendentry{EviSearch, random}
  \addplot[thick, orange!80!black, dashdotted]
    coordinates {(0,92.27) (100,100)};
  \addlegendentry{Search Agent, random}
  \addplot[thin, black!60, densely dotted, domain=0:100, samples=2] {93.21};
  \addlegendentry{Oracle selection}
  \node[font=\scriptsize, anchor=south west, inner sep=1pt] at (axis cs:16.5,95.3) {95.2};
  \node[font=\scriptsize, anchor=north west, inner sep=1pt, orange!80!black] at (axis cs:16.5,93.35) {93.5};
\end{axis}
\end{tikzpicture}

%% file: sections/demo.tex
\section{Demonstration}
\label{sec:demo}

The demonstration exposes the complete workflow of a table owner through a web interface
(Appendix~\ref{app:ui}); a publicly accessible deployment serves the publications of the evaluation with their
extraction runs cached, and the system can be installed locally from the released repository. In this deployment the
authors act as the table's reviewers: they curate the column definitions, review the cells the system flags, and
refine the schema and the knowledge notes from those reviews.

\paragraph{Defining a table.} The user provides the column headers of the target table and one example row. The
drafting agent proposes a definition for every column and attaches clarifying questions to those whose intent is
ambiguous. The user accepts, edits, or answers each definition and locks the schema as a version.

\paragraph{Extracting and inspecting.} Uploaded publications are extracted column group by column group. Selecting
a cell opens the source PDF at the cited page with the value itself highlighted, or, for a derived value, the numbers
it was computed from, alongside the three candidates, the Reconciliation Agent's verdicts, and whether the verifier's
independent reading reproduced the value.

\paragraph{Reviewing.} The review queue presents the cells selected by disagreement-based triage, on average
\queueFrac\% of the cells of a table. Each entry displays the competing candidates and the verifier's reading, so
that most cells can be resolved without navigating the source document.

\paragraph{Correcting.} A correction is stored with the reviewer's rationale. Corrections that reflect an
interpretation of the schema are proposed as edits to the governing knowledge note; the knowledge page then records
the edit, its author, and the fingerprint of the knowledge base, with the runs that used each version. Re-extracting
a publication applies the revised knowledge.

%% file: sections/conclusion.tex
\section{Conclusion}
\label{sec:conclusion}

EviSearch demonstrates that automated evidence extraction can be made traceable, private, and responsive to its
users. Every value it produces is attributed and independently verified, the
disagreement between its agents concentrates the effort of human review where errors are most likely, and reviewer
feedback aligns its behaviour through the schema and a curation knowledge base rather than through retraining. We
release the system to support trustworthy, human-supervised automation of living evidence synthesis.

%% file: sections/limitations.tex
\section*{Limitations}

The evaluation is confined to a single disease area, and generalization to other clinical
domains and table schemas remains to be established. Accuracy is measured by an LLM judge (Claude Opus~5) against a written rubric
rather than by expert adjudication of every cell. Reviewer effort is estimated under the assumption that a reviewer
corrects every erroneous cell they inspect, which bounds rather than measures the accuracy attainable in practice;
likewise, schema review was performed by a simulated table owner rather than by clinicians. The autonomous accuracy of
the full system does not exceed that of its strongest individual agent, and its advantage lies in attribution and
triage. Values that are not printed on their cited page, because they were computed without stated arithmetic or read from
a figure, are shown at the cited page without a highlighted region. Finally, layout parsing relies on an external service
invoked once per document at ingestion; all subsequent model inference uses open-weight models that can be served locally.

%% file: sections/ethics.tex
\section*{Ethics Statement}

EviSearch processes published, peer-reviewed trial reports that contain no personally identifiable patient
information. All language-model inference is performed by open-weight models on local hardware, so documents are not
transmitted to external model providers during extraction; layout parsing is the single external operation and occurs
once per document at ingestion. Every extracted value is stored with its attribution to support verification and
correction, and reviewer edits are logged with their rationale. EviSearch is intended to assist, not replace, expert
curation: its outputs should be verified before they inform meta-analyses or clinical decisions.

%% file: sections/appendix.tex
\section{Evaluation Protocol}
\label{app:eval}

\paragraph{Rubric.} Columns fall into three categories, each with its own scoring instructions: \emph{exact match}
(identifiers, trial names, categorical labels), \emph{numeric tolerance} (counts, percentages, medians, rates), and
\emph{structured text} (regimens, arm descriptions, endpoint lists). The judge rates correctness and completeness
separately on $\{0, 0.5, 1\}$. Seven conventions, fixed before any system was scored, resolve recurring ambiguities
uniformly across systems:
(C1)~counts must match exactly, percentages within one percentage point or two percent relative, other quantities
within 0.1 or at a different rounding;
(C2)~a required number counts as present only if it is within tolerance, and per-arm values do not substitute for a
required pooled total;
(C3)~``not reached'' and ``not estimable'' are substantive values, distinct from an absent value;
(C4)~a category label implied by the components a prediction names counts as present;
(C5)~parenthetical context in a reference value is optional for exact-match columns;
(C6)~in yes/no columns, ``No'' and an empty value are equivalent;
(C7)~where a definition requests a ``count and/or percentage'', either suffices, whereas ``N~(\%)'' requires both.

\paragraph{Blind judging.} Each item presented to the judge contains only the column, its definition, its
category, the reference value, and the predicted value; the producing system is never disclosed, and identical
predictions from different systems are judged once. Judgements are cached by document, column, and normalized
prediction, so every system is scored against the same labels. Items on which a first judgement was uncertain were
adjudicated a second time in isolation.

\paragraph{Reference annotations.} Scores are computed against the reference annotations as released. An audit of
the reference values against their source pages found that \refAuditShare\% of reference values contradict their cited page;
these were not corrected for any reported result.

\section{Verifier Readings and Admission}
\label{app:ablation}

The attribution verifier is instructed to formulate its own answer from the cited pages before comparing it with a
candidate. Its reading can therefore be used in two ways: as a condition for admission, or as information for the adjudicating agent
and the reviewer. Table~\ref{tab:admission} reports, for each use, how the verifier's negative readings affected the
cells they concerned, over two runs.

\begin{table}[h]
\centering
\small
\setlength{\tabcolsep}{3pt}
\begin{tabular}{@{}p{3.9cm}cc@{}}
\toprule
 & \textbf{Agreement-} & \textbf{Provenance-} \\
 & \textbf{gated} & \textbf{gated} \\
\midrule
Values dropped after a negative reading & \agreeDropped & \provDropped \\
\quad of which correct (\% of dropped) & \agreeDropCorrect & \provDropCorrect \\
Cells corrected by the reading & \agreeFixed & \provFixed \\
Cells made incorrect by the reading & \agreeBroke & \provBroke \\
Corroboration mark informative & no & yes \\
\bottomrule
\end{tabular}
\caption{Effect of the verifier's negative readings under the two admission policies, as a percentage of all
cells unless stated; two runs pooled. Under
agreement gating a negative reading excludes the value; under provenance gating it informs the adjudicating agent,
and a value it does not reproduce is admitted, marked, and routed to review.}
\label{tab:admission}
\end{table}

Under agreement gating, a value the verifier did not reproduce could not be admitted, so the verifier's failures to
locate a value became omissions in the table. Of the values it caused to be dropped, more were correct than incorrect,
and its readings made more cells incorrect than they corrected. Under provenance gating, the same readings inform
the Reconciliation Agent without binding it; they correct twice as many cells while making no more cells incorrect,
and the corroboration mark distinguishes reproduced from unreproduced values rather than holding for every admitted
value by construction.

\section{Agent Instructions}
\label{app:prompts}

We summarize the instructions of the Reconciliation Agent, which carries the system's adjudication logic; the complete
instructions of all agents are released with the code.

\paragraph{Independent extraction phase.} \emph{``You extract clinical trial values from a research paper. You are
given columns, each with a definition. Fill every column.''} The agent is directed to read the opening pages first,
to retrieve pages for the columns that remain open, to open pages with their images when a value may be printed in a
figure or in a table rendered as an image, and to confirm located values with the attribution verifier. It must copy
values as printed, give the page and the supporting text for each value, never report a value the publication does not
state or read one from a plotted curve, and either answer every column or list the pages it examined.

\paragraph{Adjudication phase.} \emph{``You review two independent extractions of the same paper and decide each
column's final value.''} The agent submits unanimous values directly and, where the candidates differ, reads the pages
they cite. It is instructed that the verifier is a second reader confined to the pages it is given, so that a
disagreeing reading is a discrepancy to resolve by inspection rather than a ruling, and that the verifier's failure to
find a value is not evidence of its absence. Answers that are compatible at different levels of detail are merged;
an abstention requires the pages read and what they report; and every candidate, including the agent's own, receives
a verdict.

\section{Interface}
\label{app:ui}

\begin{figure*}[t]
\centering
\includegraphics[width=\textwidth]{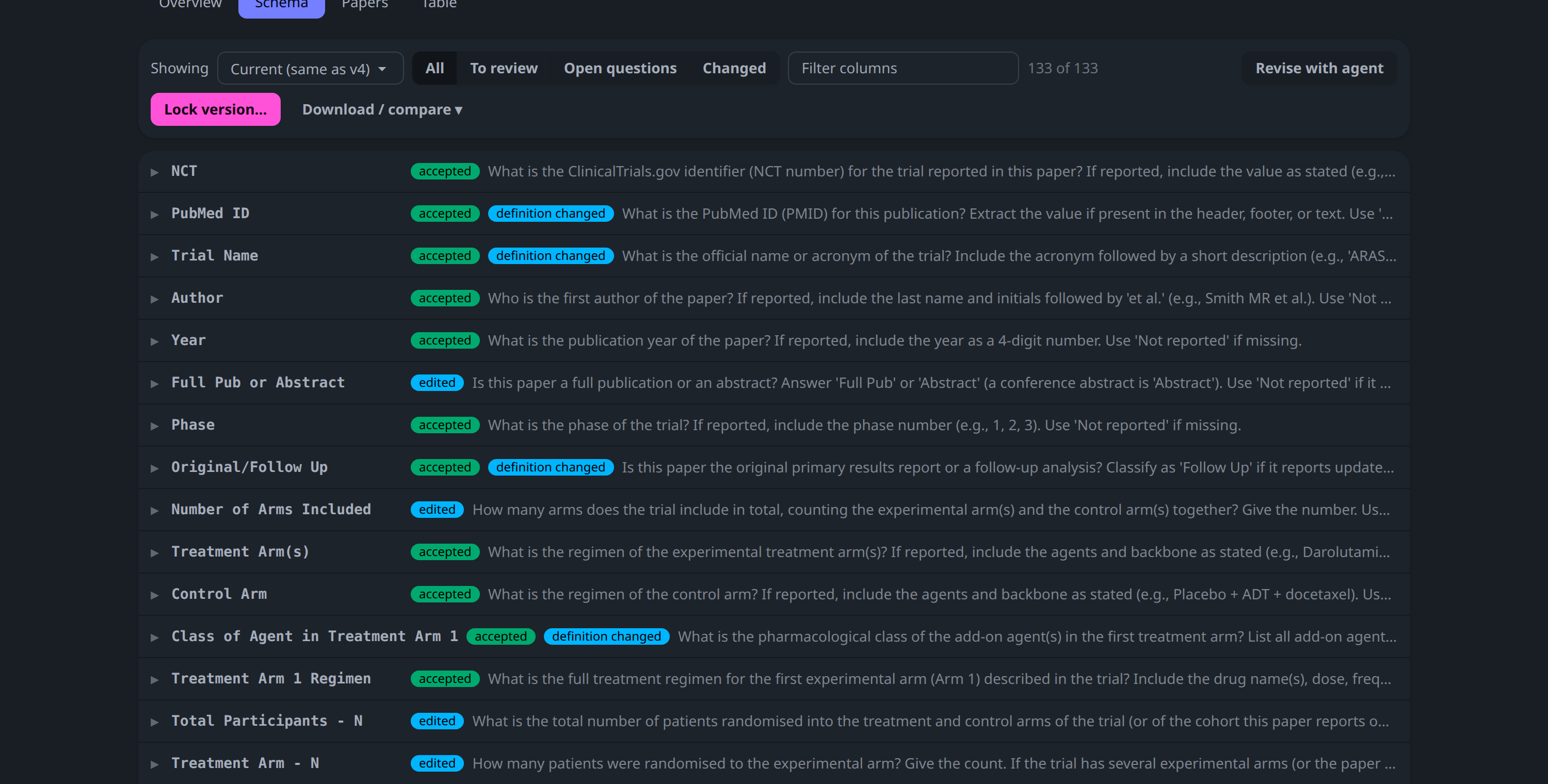}
\caption{Defining a table. Each drafted definition carries its status: accepted as drafted, edited by the owner, or
changed since the previous version. The version is locked before extraction.}
\label{fig:ui-schema}
\end{figure*}

\begin{figure*}[t]
\centering
\includegraphics[width=\textwidth]{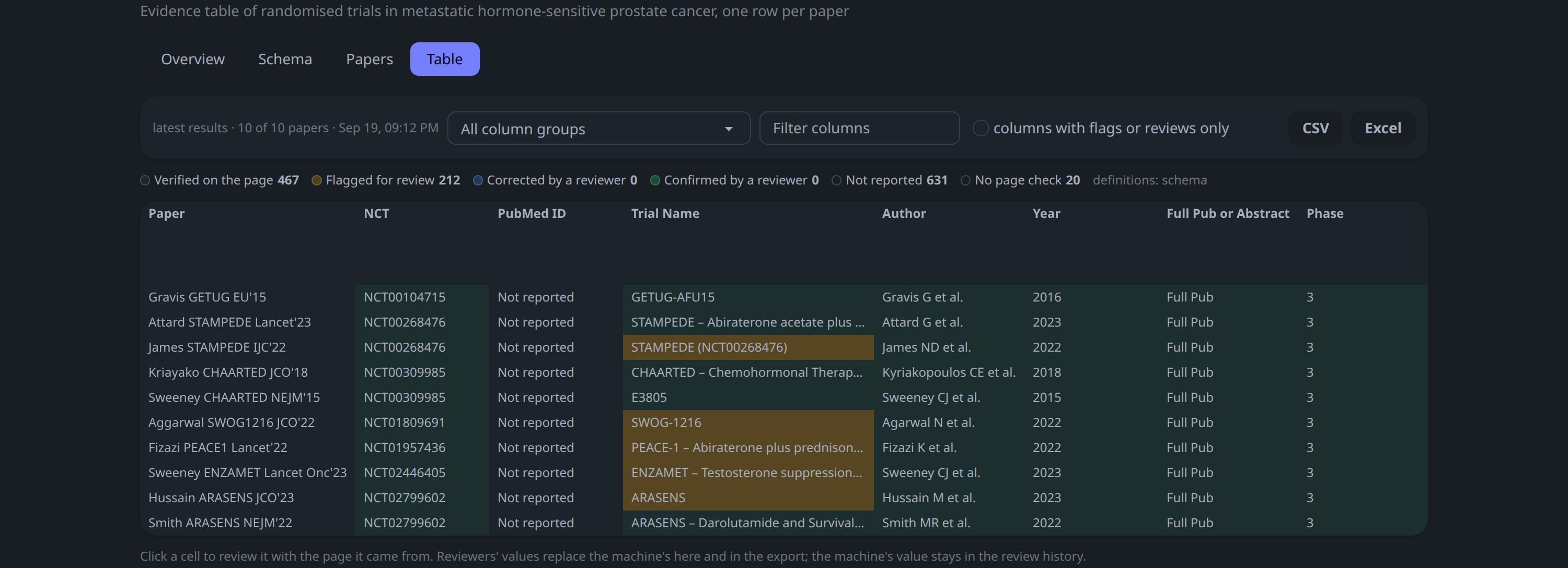}
\caption{The extracted table. Every cell carries its state, verified on the cited page, flagged for review,
corrected or confirmed by a reviewer, or not reported, and opens at its attribution when selected.}
\label{fig:ui-table}
\end{figure*}

\begin{figure*}[t]
\centering
\includegraphics[width=\textwidth]{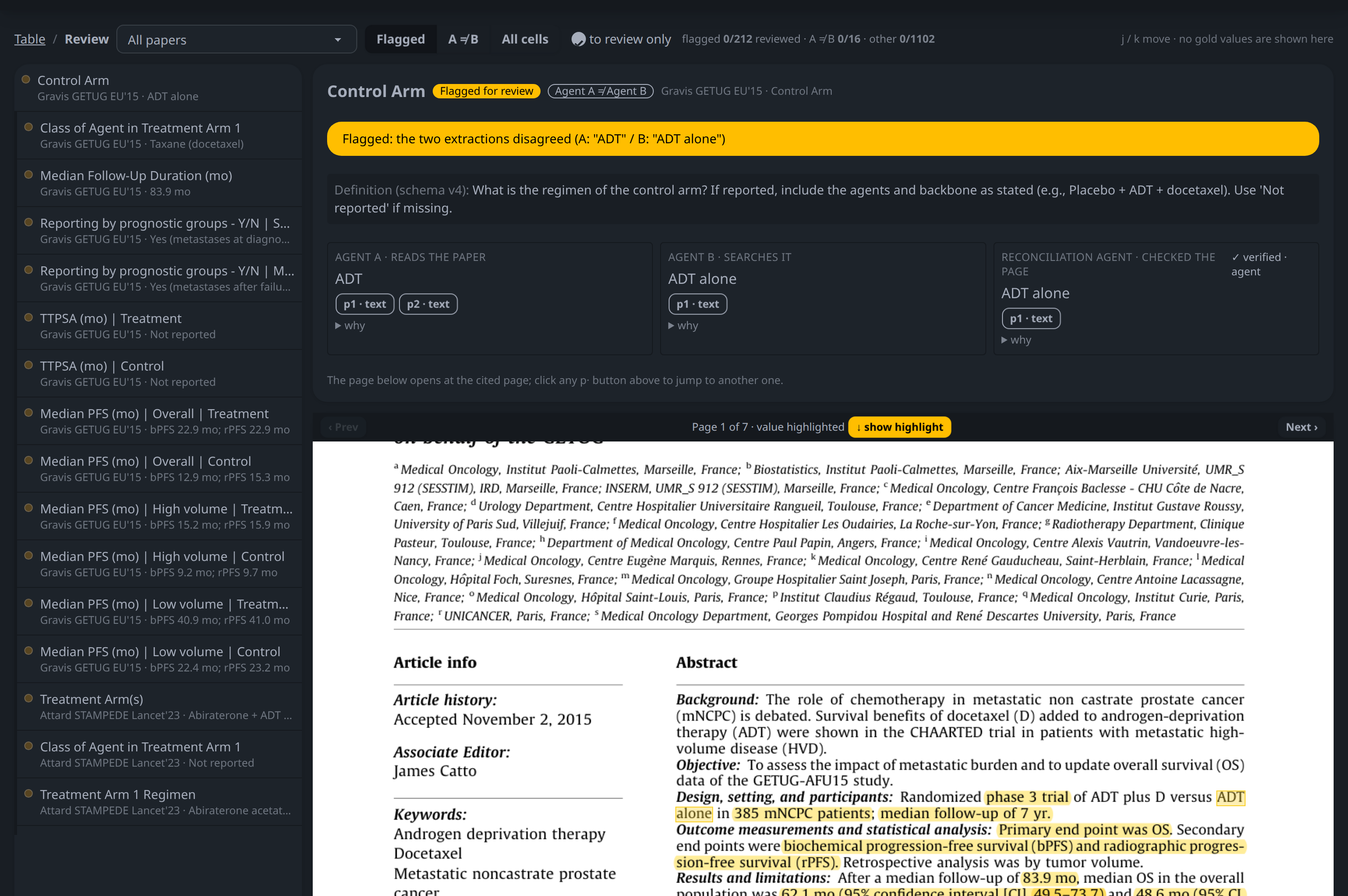}
\caption{The review view for one queued cell. The extraction agents disagree (``ADT'' against ``ADT alone''); the
three candidates are shown with their cited pages and the verifier's mark, and the source page opens with the value
highlighted.}
\label{fig:ui-review}
\end{figure*}

\begin{figure*}[t]
\centering
\includegraphics[width=\textwidth]{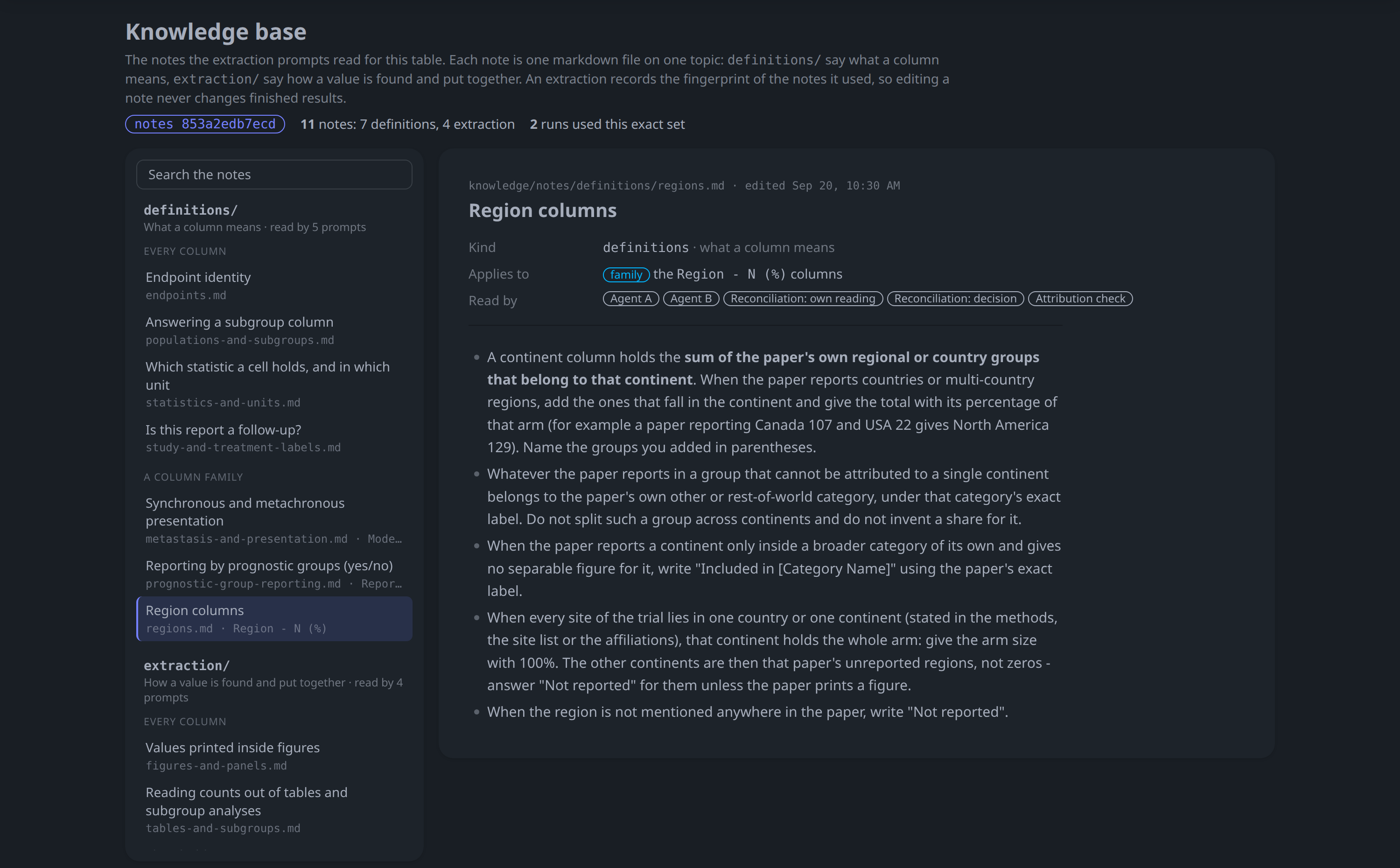}
\caption{The curation knowledge base. Each note states its scope, the agents that receive it, the conventions it
replaced, and the reason for its last edit; the fingerprint identifies the version that extraction runs recorded.}
\label{fig:ui-knowledge}
\end{figure*}

Figures~\ref{fig:ui-schema}--\ref{fig:ui-knowledge} show the stages of the workflow of Section~\ref{sec:demo} in the
deployed interface.

